\documentclass[11pt]{article}
\usepackage{arxiv}
\usepackage[T1]{fontenc}
\usepackage[utf8]{inputenc}
\usepackage{lmodern}
\usepackage{microtype}
\usepackage{amsmath,amssymb}
\usepackage{booktabs}
\usepackage{tabularx}
\usepackage{multirow}
\usepackage{graphicx}
\usepackage{caption}
\usepackage{subcaption}
\usepackage{enumitem}
\usepackage{siunitx}
\usepackage{url}
\usepackage[hidelinks,pdftitle={AI World Cup 2026: Benchmarking Large Language Models for End-to-End Football Tournament Prediction},pdfauthor={Jonaid Shianifar and Iias Faiud},pdfkeywords={large language models, forecasting, sports analytics, World Cup, reproducibility}]{hyperref}
\usepackage[round,authoryear]{natbib}
\usepackage{xcolor}
\usepackage{listings}
\usepackage{float}

\setlist[itemize]{leftmargin=1.5em,itemsep=2pt,topsep=3pt}
\setlist[enumerate]{leftmargin=1.7em,itemsep=2pt,topsep=3pt}
\newcommand{\commitid}{\texttt{f83ea90}}

\title{AI World Cup 2026: Benchmarking Large Language Models for End-to-End Football Tournament Prediction}
\author{
Jonaid Shianifar \quad Iias Faiud\\
\small AI World Cup Project
}
\date{31 July 2026}

\begin{document}
\maketitle

\begin{abstract}
Large language models (LLMs) are now regularly asked to forecast real-world events, but comparisons are often difficult because models receive different information, use different tools, and are evaluated under different rules. This paper reports the completed \emph{AI World Cup} benchmark, in which ten LLM-based assistants made a single pre-tournament forecast of the entire 2026 FIFA World Cup. Every submission used the same tournament snapshot, prompt, JSON schema, and scoring procedure. The forecasts covered group-stage scores, group rankings, the knockout bracket, final placings, confidence values, and short explanations. After all 104 matches had been played, GPT-5.5 Thinking finished first with 744 points, followed by GPT-5.5 with 717, Gemini with 699, and Qwen 3.7 with 687. GPT-5.5 Thinking was also the only model to select Spain, which defeated Argentina 1--0 in the final, as champion. The final ranking was driven mainly by knockout performance: total score was strongly correlated with knockout points ($r=0.986$), but showed little relationship with group-stage match points ($r=0.055$), group-standing points ($r=-0.103$), or their combined pre-knockout score ($r=-0.054$). Match-level accuracy produced a different ordering. Claude Sonnet 4.6 correctly predicted the largest number of group-stage outcomes (63.89\%) but placed sixth overall. Average self-reported confidence was also unrelated to either outcome accuracy ($r=-0.060$) or total score ($r=-0.067$). The results suggest that forecasting a complete tournament tests something different from predicting matches one at a time, while also showing how strongly a bracket-based leaderboard can depend on scoring design. The benchmark materials, raw responses, and scoring code are released to support replication and future extensions.
\end{abstract}


\section{Introduction}
Forecasting is a useful way to test how a model reasons when the correct answer is not yet known. It requires the model to work with incomplete evidence, weigh conflicting signals, and commit to a prediction before the event takes place. This makes prospective forecasting different from many static question-answering benchmarks, where answers may already appear in training data. Once the event is resolved, the prediction can be evaluated directly and transparently \citep{tetlock2014superforecasting}.

A football tournament is a particularly demanding forecasting problem. Individual matches are noisy, draws are common, and a single unexpected result can change the rest of the bracket. Predicting the whole competition therefore involves more than choosing the likely winner of each match. The forecaster must also produce a consistent path from the group stage to the final. With 48 teams and 104 fixtures, the 2026 FIFA World Cup provides a large test of both match-level judgement and long-horizon tournament reasoning.

The \emph{AI World Cup} project was created as a public and reproducible comparison of LLM assistants.\footnote{Project repository: \url{https://github.com/jonaidshianifar/ai-world-cup}} Each model received the same generated prompt and tournament data, and each response had to follow a common JSON schema. The submissions were then evaluated by one open-source scoring pipeline. Predictions were collected manually through the available consumer interfaces rather than through paid APIs. This made it possible to include widely accessible assistants while still preserving the original responses for later inspection.

This paper makes the following contributions:
\begin{itemize}
    \item a completed full-tournament evaluation of ten LLM assistants using a common prompt, snapshot, response schema, and transparent scoring implementation;
    \item an independent audit of the final leaderboard, including stage decomposition, accuracy counts, confidence analysis, and result provenance;
    \item evidence that knockout-bracket performance, rather than group-stage match prediction, determined the composite ranking;
    \item a comparison with three contemporaneous 2026 World Cup LLM benchmarks; and
    \item a reproducibility and evaluation roadmap for a stronger second version of the benchmark.
\end{itemize}

\section{Related Work}
\subsection{Prospective LLM forecasting}
Static benchmarks can be affected by data contamination or by repeated prompt tuning against known answers. Prospective forecasting reduces this risk because predictions are recorded before the event is resolved. When models provide probabilities, strictly proper scoring rules such as the Brier score or logarithmic score are generally preferred because they reward honest uncertainty estimates \citep{brier1950verification,gneiting2007strictly}. AI World Cup began instead as a public points-based competition that combined exact scores, match outcomes, group rankings, and bracket progression. This format is easy to understand, but the resulting leaderboard inevitably depends on how those components are weighted.

\subsection{LLM football forecasting benchmarks}
Three independent projects evaluated LLMs during the same tournament. \citet{wang2026worldcuparena} introduced WorldCupArena, a dynamic benchmark covering 104 matches and 13 systems. It evaluated results, exact scores, near-score quality, player events, and match statistics, and compared systems with human and betting-market baselines. \citet{ding2026wcagents} introduced WC2026-Agents, in which four web-enabled agents used a shared search--act--reflect loop to output 1X2 probabilities and virtual bets for every match; none outperformed the betting market on Brier score. \citet{schroeder2026soccerarena} introduced LLM-SoccerArena, a prospective factorial benchmark varying model, information access, prompting strategy, and forecast horizon across seven LLMs and 104 matches.

AI World Cup differs from these studies in three practical respects. It asks for one complete prediction before the tournament begins, includes consumer-facing assistants through manual submission, and scores the tournament path as well as individual matches. This design gives more weight to long-range consistency, but it also means that one early bracket error can affect many later predictions.

\section{Benchmark Design}
\subsection{Design principles}
The project specifies eight methodological principles: the same prompt, the same data snapshot, manual submission, preservation of raw responses, structured validation, gradual evaluation, a transparent leaderboard, and separation of search-enabled systems where possible. The intended comparison is therefore controlled at the prompt and input level, while remaining accessible to models offered through consumer interfaces.

\subsection{Participants}
The final populated leaderboard contains ten submissions from seven providers: GPT-5.5 Thinking, GPT-5.5, Qwen 3.7, Gemini, DeepSeek, Claude Sonnet 4.6, Mistral Medium 3.5, Perplexity, Perplexity Pro, and Grok. The labels are retained exactly as recorded in the project. They should be interpreted as submitted assistant configurations, not as timeless model families, because consumer products and serving configurations can change.

\subsection{Prompt and outputs}
A single generated prompt asked each model to return one JSON object containing:
\begin{itemize}
    \item score, outcome, winner, confidence, and short reasoning for every group-stage match;
    \item a predicted ranking for every group;
    \item a complete knockout-stage bracket;
    \item champion, runner-up, third-place, and fourth-place predictions; and
    \item individual award predictions.
\end{itemize}
Raw responses were saved before parsing. The importer then checked the expected structure and converted valid responses into structured predictions for scoring.

\subsection{Scoring system}
The benchmark uses an additive points system with three components: group-stage match predictions, predicted group standings, and knockout/tournament progression. All applicable bonuses are cumulative. Consequently, a single prediction may receive several awards when it satisfies several criteria simultaneously.

\subsubsection{Group-stage match points}
For each of the 72 group-stage matches, let $\hat{h}_m$ and $\hat{a}_m$ denote the predicted home and away goals, and let $h_m$ and $a_m$ denote the corresponding realised goals. The score for match $m$ is
\begin{equation}
S_m = 5I_{\mathrm{exact}} + 3I_{\mathrm{outcome}} + 2I_{\mathrm{winner}} + I_{\mathrm{GD}},
\label{eq:matchscore}
\end{equation}
where
\begin{align}
I_{\mathrm{exact}} &= \mathbb{I}(\hat{h}_m=h_m \ \land\ \hat{a}_m=a_m),\\
I_{\mathrm{outcome}} &= \mathbb{I}(\operatorname{sgn}(\hat{h}_m-\hat{a}_m)=\operatorname{sgn}(h_m-a_m)),\\
I_{\mathrm{GD}} &= \mathbb{I}(\hat{h}_m-\hat{a}_m=h_m-a_m).
\end{align}
For a non-draw, $I_{\mathrm{winner}}=1$ when the predicted winning team is the actual winning team. For a draw, the separate winner bonus is not awarded because the correct draw is already represented by $I_{\mathrm{outcome}}$. The complete point allocation is reported in Appendix Table~\ref{tab:matchscoring}.

The group-stage match component is therefore
\begin{equation}
S_{\mathrm{group\ matches}}=\sum_{m=1}^{72}S_m.
\end{equation}
For example, if a model predicts Spain 2--1 Argentina and the realised result is Spain 2--1 Argentina, it receives $5+3+2+1=11$ points. If it predicts Spain 1--0 Argentina, it receives $3+2+1=6$ points because the winner, outcome, and one-goal margin are correct, but the exact score is not. For an exact draw, the maximum is $5+3+1=9$ points because no separate winner bonus is used.

\subsubsection{Group-standing points}
After all matches in a group are complete, the submitted ranking is compared with the realised ranking and qualification set. For each group $g$, the score is
\begin{equation}
S_g^{\mathrm{standing}} = 5I_{\mathrm{winner}}^{(g)} + 5I_{\mathrm{top2}}^{(g)}
+3N_{\mathrm{qual}}^{(g)}+2N_{\mathrm{rank}}^{(g)},
\label{eq:standingscore}
\end{equation}
where $I_{\mathrm{winner}}^{(g)}$ indicates that the group winner is correct, $I_{\mathrm{top2}}^{(g)}$ indicates that the unordered set of the first two teams is correct, $N_{\mathrm{qual}}^{(g)}$ is the number of correctly predicted qualifiers from that group, and $N_{\mathrm{rank}}^{(g)}$ is the number of teams placed in their exact final position. The awards are cumulative; the complete allocation is reported in Appendix Table~\ref{tab:standingscoring}.

The total group-standing score is
\begin{equation}
S_{\mathrm{group\ standings}}=\sum_{g=1}^{12}S_g^{\mathrm{standing}}.
\end{equation}
For example, if the predicted top two are exactly the correct two teams in the correct order, the prediction receives the group-winner bonus, the top-two-set bonus, the qualifier bonuses for those teams, and their exact-rank bonuses. Additional points are awarded when the predicted third- and fourth-placed teams also occupy their exact positions and when a third-placed team qualifies.

\subsubsection{Knockout and final-ranking points}
Knockout scoring evaluates whether each predicted team reaches the corresponding tournament stage. Let $C_r$ denote the number of teams correctly predicted to reach round $r$. Stage awards accumulate: a correctly predicted finalist can also earn points for being correctly predicted in each preceding knockout round. The progression score is
\begin{align}
S_{\mathrm{progression}}={}&2C_{\mathrm{R32}}+4C_{\mathrm{R16}}+6C_{\mathrm{QF}} \\
&+8C_{\mathrm{SF}}+12C_{\mathrm{Final}}.
\label{eq:progressionscore}
\end{align}
Final placings are then scored separately:
\begin{equation}
S_{\mathrm{placing}}=20I_{\mathrm{champion}}+10I_{\mathrm{runner\text{-}up}}
+8I_{\mathrm{third}}+5I_{\mathrm{fourth}}.
\label{eq:placingscore}
\end{equation}
The total knockout score is
\begin{equation}
S_{\mathrm{knockout}}=S_{\mathrm{progression}}+S_{\mathrm{placing}}.
\end{equation}

The stage-by-stage point allocation is summarised in Appendix Table~\ref{tab:knockoutscoring}.

For example, a team correctly predicted to reach the final can contribute $2+4+6+8+12=32$ progression points when its entire path is represented correctly in the stage sets. If that team is also the correctly predicted champion, it earns a further 20 points. The scoring is based on correctly predicted stage membership and final position, not on exact knockout scorelines.

\subsubsection{Total benchmark score}
Models are ranked using the total score
\begin{equation}
S_{\mathrm{total}} = S_{\mathrm{group\ matches}} + S_{\mathrm{group\ standings}} + S_{\mathrm{knockout}}.
\label{eq:totalscore}
\end{equation}
In the completed results, group-standing points were evaluated for all ten submissions and ranged from 221 to 264, making this component a material part of the final score.

\subsection{Evaluation measures}
For each model, we report the overall score, its three components, exact-score accuracy, outcome accuracy, average confidence, and predicted champion. The accuracy fields change in steps of $1/72$, which confirms that they refer to the 72 group-stage matches; we therefore also report the corresponding numbers of correct predictions. Pearson and Spearman correlations are used to describe relationships between the score components. With only ten submissions, these correlations should be read as descriptive summaries rather than as estimates that generalise to all LLMs.

\section{Final Result Audit and Provenance}
\subsection{Data freeze}
The analysis uses the results stored in repository commit \commitid, created on 31 July 2026 when actual group standings were reconstructed and the group-standing rubric was applied. The corresponding website export records ten models, 1,040 predictions, and all 104 fixtures, with a generation timestamp of 31 July 2026 at 14:10 UTC. The final leaderboard identifies GPT-5.5 Thinking as the only submission selecting Spain as champion; the recorded final was Spain 1--0 Argentina.

\subsection{Independent checks}
We verified that (i) total points equal group-stage match plus group-standing plus knockout points for every entry; (ii) ranks are monotonically ordered by total score; (iii) accuracy fractions map to integer counts over 72 group-stage matches; (iv) the public JSON contains ten models and 1,040 model--fixture predictions; and (v) the recorded champion prediction of the winner matches the recorded final result. All descriptive statistics and correlations were recomputed directly from the archived JSON values rather than copied from the website presentation.

\section{Results}
\subsection{Overall tournament performance}
Table~\ref{tab:leaderboard} reports the final tournament scores. GPT-5.5 Thinking ranked first with 744 points, followed by GPT-5.5 with 717, Gemini with 699, and Qwen 3.7 with 687. The remaining six models scored between 496 and 599 points.

The component scores show that no model was uniformly strongest across every part of the benchmark. Qwen 3.7 and Perplexity Pro shared the highest group-stage match score (270), while GPT-5.5 achieved the highest group-standing score (264). GPT-5.5 Thinking led the knockout component by a clear margin, scoring 242 points compared with 196 for Qwen 3.7 and 192 for both GPT-5.5 and Gemini. This knockout advantage was sufficient to overcome its lower score in the two group-phase components.

\begin{table}[H]
\centering
\caption{Final AI World Cup 2026 leaderboard frozen to commit \commitid. GS: group-stage match points; ST: group-standing points; KO: knockout points; OA: outcome accuracy; ESA: exact-score accuracy.}
\label{tab:leaderboard}
\resizebox{\textwidth}{!}{%
\begin{tabular}{rllrrrrrrrl}
\toprule
Rank & Model & Provider & Total & GS & ST & KO & OA (\%) & ESA (\%) & Champion \\
\midrule
1 & GPT-5.5 Thinking & OpenAI & \textbf{744} & 254 & 248 & \textbf{242} & 58.33 & 11.11 & Spain \\
2 & GPT-5.5 & OpenAI & 717 & 261 & \textbf{264} & 192 & 62.50 & 8.33 & Brazil \\
3 & Gemini & Google & 699 & 259 & 248 & 192 & 61.11 & 8.33 & France \\
4 & Qwen 3.7 & Qwen & 687 & \textbf{270} & 221 & 196 & 58.33 & \textbf{13.89} & Argentina \\
5 & DeepSeek & DeepSeek & 599 & 247 & 246 & 106 & 58.33 & 8.33 & Brazil \\
6 & Claude Sonnet 4.6 & Anthropic & 591 & 268 & 255 & 68 & \textbf{63.89} & 8.33 & Argentina \\
7 & Mistral Medium 3.5 & Mistral AI & 568 & 246 & 248 & 74 & 59.72 & 6.94 & France \\
8 & Perplexity & Perplexity & 552 & 257 & 257 & 38 & 58.33 & 11.11 & Brazil \\
9 & Perplexity Pro & Perplexity & 530 & \textbf{270} & 260 & 0 & 58.33 & \textbf{13.89} & Brazil \\
10 & Grok & xAI & 496 & 257 & 239 & 0 & 58.33 & 11.11 & France \\
\bottomrule
\end{tabular}}
\end{table}

\subsection{Where the leaderboard separation came from}
\begin{figure}[t]
\centering
\includegraphics[width=0.92\textwidth]{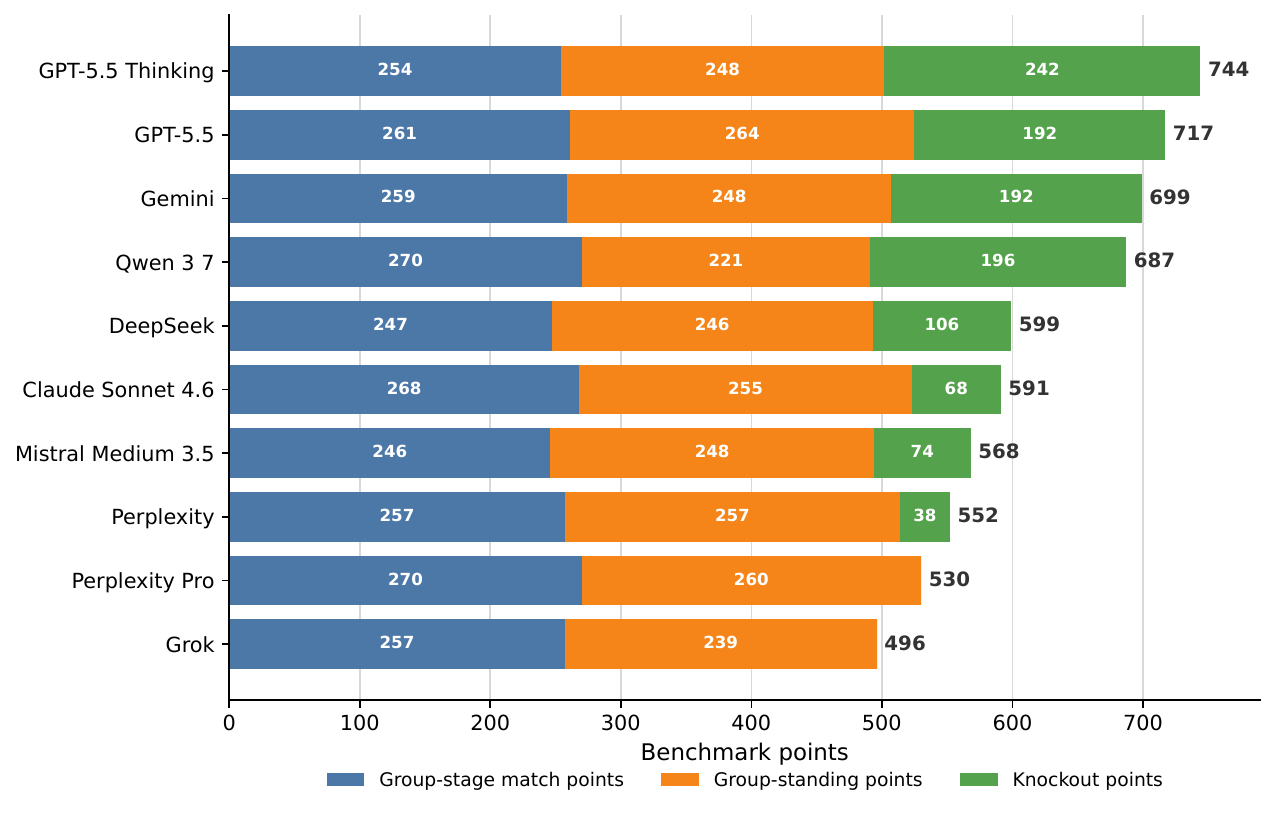}
\caption{Final score composition. Group-stage match and group-standing points are relatively similar across models, whereas knockout points create most of the separation in total score.}
\label{fig:composition}
\end{figure}

Figure~\ref{fig:composition} decomposes each model's total into group-stage match, group-standing, and knockout points. The two group-phase components were relatively concentrated: group-stage match scores ranged from 246 to 270, and group-standing scores ranged from 221 to 264. Knockout scores, by contrast, ranged from 0 to 242. Their sample standard deviations were 8.65, 12.20, and 88.65 points, respectively.

This difference in dispersion is reflected in the correlations with the final total. Total points were strongly associated with knockout points ($r=0.986$; Spearman $\rho=0.945$), but showed little relationship with group-stage match points ($r=0.055$), group-standing points ($r=-0.103$), or their combined score ($r=-0.054$). These correlations are descriptive because they are based on only ten submissions, but the contrast is large enough to show that knockout forecasting produced most of the final rank separation.

\begin{figure}[t]
\centering
\includegraphics[width=0.72\textwidth]{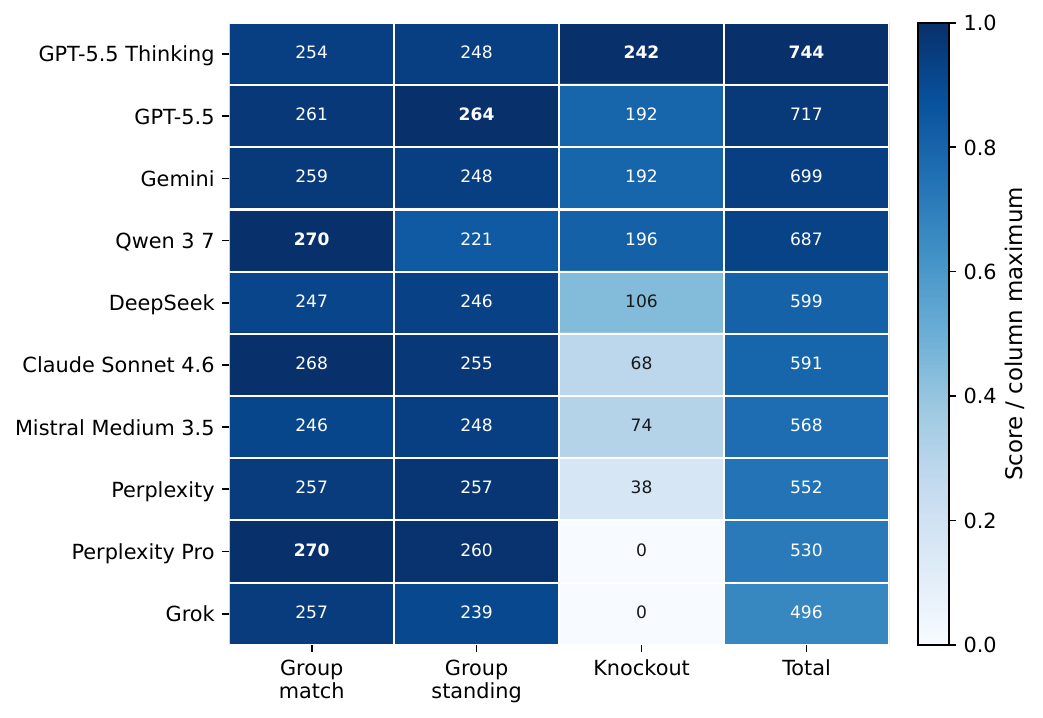}
\caption{Normalised score profile by model. Each component is scaled relative to the highest observed score in that column, with raw point values shown inside the cells.}
\label{fig:heatmap}
\end{figure}

\begin{figure}[t]
\centering
\includegraphics[width=0.90\textwidth]{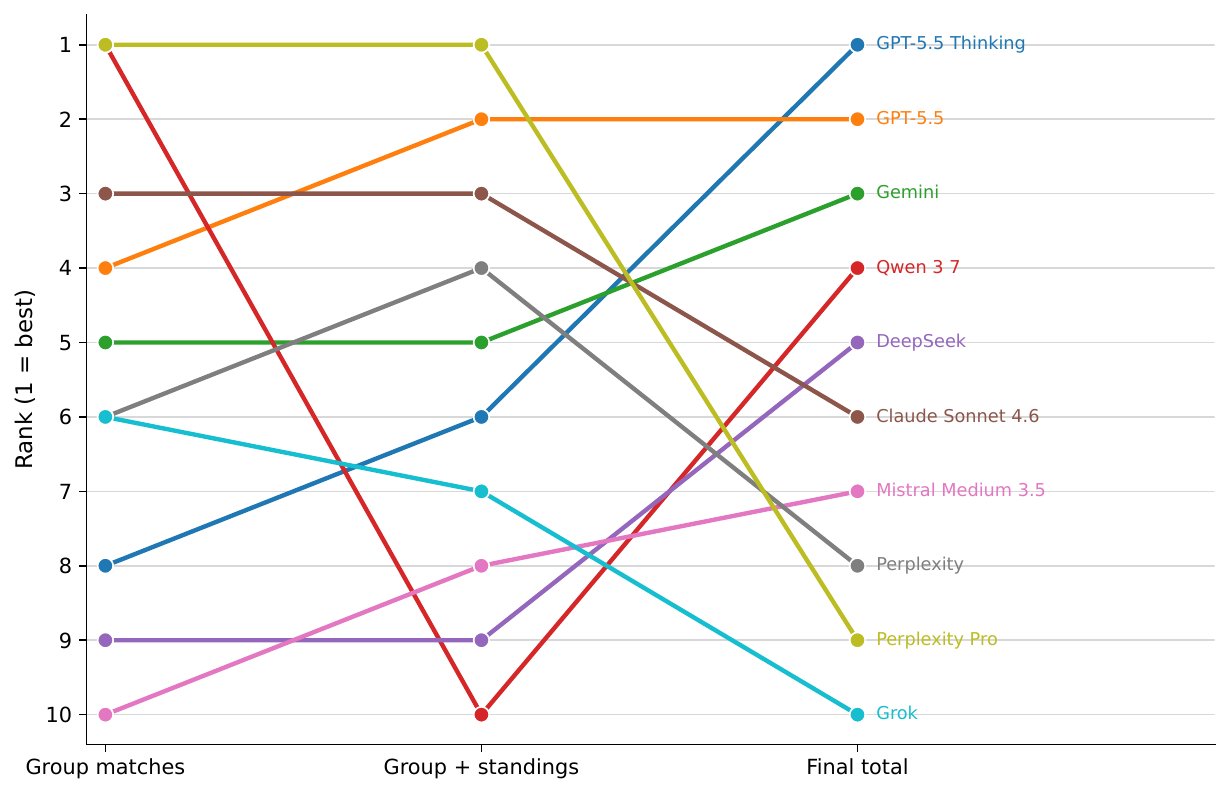}
\caption{Rank evolution across evaluation stages. Ranks are calculated after group-stage match scoring, after adding group-standing points, and after adding knockout points. Lower values indicate better ranks.}
\label{fig:rankevolution}
\end{figure}

The normalised profiles in Figure~\ref{fig:heatmap} make the same pattern visible across all four reported score columns. Figure~\ref{fig:rankevolution} then shows how rankings changed as additional components were included. Perplexity Pro ranked first on group-stage match points and remained first after group-standing points were added, but finished ninth because it earned no knockout points. GPT-5.5 Thinking followed the opposite trajectory: it ranked eighth on group-stage match points and sixth after the group-standing component, before moving to first once knockout performance was included.

The dominance of the knockout component reflects both genuine tournament-level forecasting and the path-dependent nature of bracket prediction. An early mistake can invalidate several later pairings, even when some of the predicted teams remain plausible. The overall score should therefore be interpreted as a measure of the realised end-to-end tournament forecast rather than as a pure measure of independent match-prediction ability.

\subsection{Match prediction and tournament prediction gave different leaders}
Table~\ref{tab:accuracycounts} expresses the group-stage accuracy rates as counts out of 72 matches. Claude Sonnet 4.6 predicted the correct outcome in 46 matches (63.89\%), the best result in this measure. GPT-5.5 followed with 45 correct outcomes, and Gemini with 44. Qwen 3.7 and Perplexity Pro recorded the most exact scorelines, with ten each. GPT-5.5 Thinking, the overall tournament winner, correctly predicted 42 outcomes and eight exact scores.

\begin{table}[H]
\centering
\caption{Group-stage prediction accuracy expressed as counts out of 72 matches.}
\label{tab:accuracycounts}
\begin{tabular}{lrrr}
\toprule
Model & Correct outcomes & Exact scores & Mean confidence (\%) \\
\midrule
GPT-5.5 Thinking & 42 & 8 & 60.68 \\
Qwen 3.7 & 42 & 10 & 72.21 \\
GPT-5.5 & 45 & 6 & 62.69 \\
Gemini & 44 & 6 & 70.24 \\
DeepSeek & 42 & 6 & 71.44 \\
Claude Sonnet 4.6 & \textbf{46} & 6 & 65.22 \\
Mistral Medium 3.5 & 43 & 5 & 73.27 \\
Perplexity & 42 & 8 & 60.64 \\
Perplexity Pro & 42 & \textbf{10} & 61.73 \\
Grok & 42 & 8 & 68.61 \\
\bottomrule
\end{tabular}
\end{table}

\begin{figure}[t]
\centering
\includegraphics[width=0.80\textwidth]{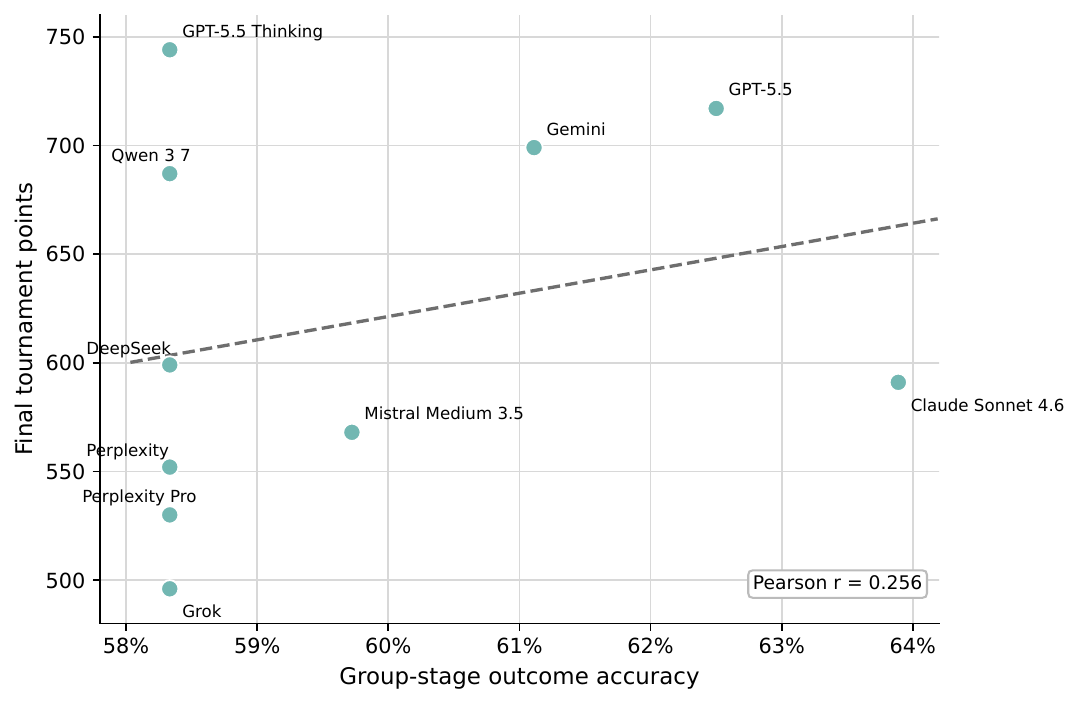}
\caption{Group-stage outcome accuracy versus final tournament score. Strong local match accuracy did not guarantee a high tournament-wide score.}
\label{fig:accuracyvstotal}
\end{figure}

\begin{figure}[t]
\centering
\includegraphics[width=0.90\textwidth]{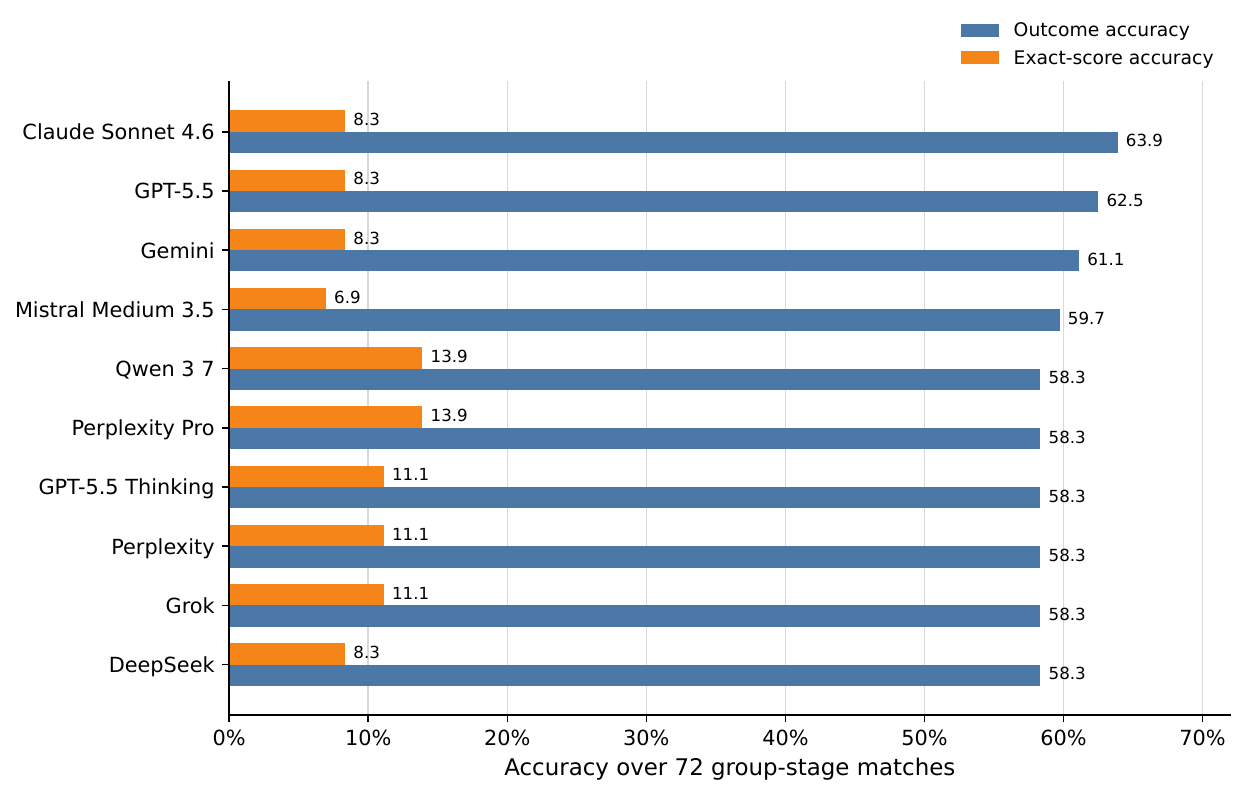}
\caption{Outcome and exact-score accuracy across the ten submissions. Exact scoreline prediction was consistently more difficult than predicting the categorical match outcome.}
\label{fig:accuracycomparison}
\end{figure}

Figures~\ref{fig:accuracyvstotal} and~\ref{fig:accuracycomparison} illustrate the distinction between local and tournament-level performance. A high outcome accuracy did not necessarily lead to a high final score, and exact-score accuracy remained much lower than outcome accuracy for every model. The two leaderboards therefore answer different questions: one rewards correct predictions over individual group-stage matches, while the other rewards a coherent forecast of the tournament as a whole.

\subsection{Champion selections and the winning margin}
\begin{figure}[t]
\centering
\includegraphics[width=0.82\textwidth]{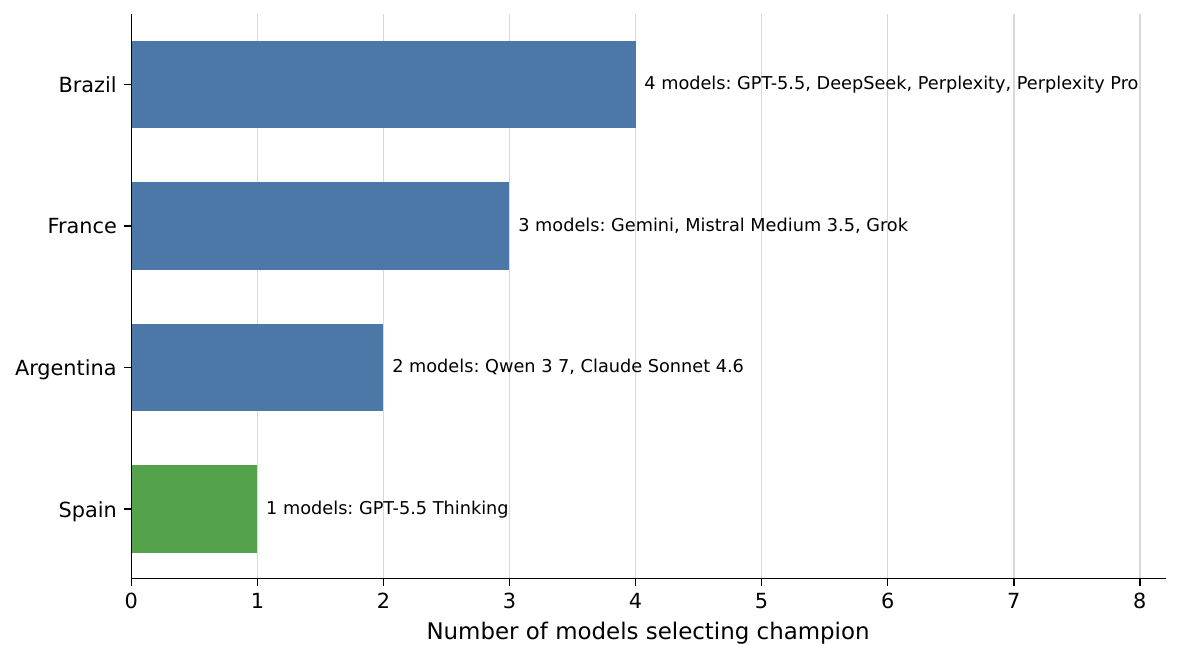}
\caption{Champion selections across the ten submissions. Spain, the eventual winner, was selected by one model.}
\label{fig:championconsensus}
\end{figure}

The ten models selected four possible champions: Brazil was chosen by four models, France by three, Argentina by two, and Spain by one. Figure~\ref{fig:championconsensus} summarises these selections. GPT-5.5 Thinking was the only model to select Spain, which ultimately defeated Argentina 1--0 in the final.

The correct champion selection was important, but it does not by itself explain the full winning margin. Before knockout points were counted, GPT-5.5 led GPT-5.5 Thinking by 23 points (525 versus 502). GPT-5.5 Thinking then earned 242 knockout points, 50 more than GPT-5.5, producing a final lead of 27 points. Its advantage therefore came from the broader predicted knockout path as well as the champion bonus.

\subsection{Self-reported confidence did not reflect accuracy}
\begin{figure}[t]
\centering
\includegraphics[width=0.82\textwidth]{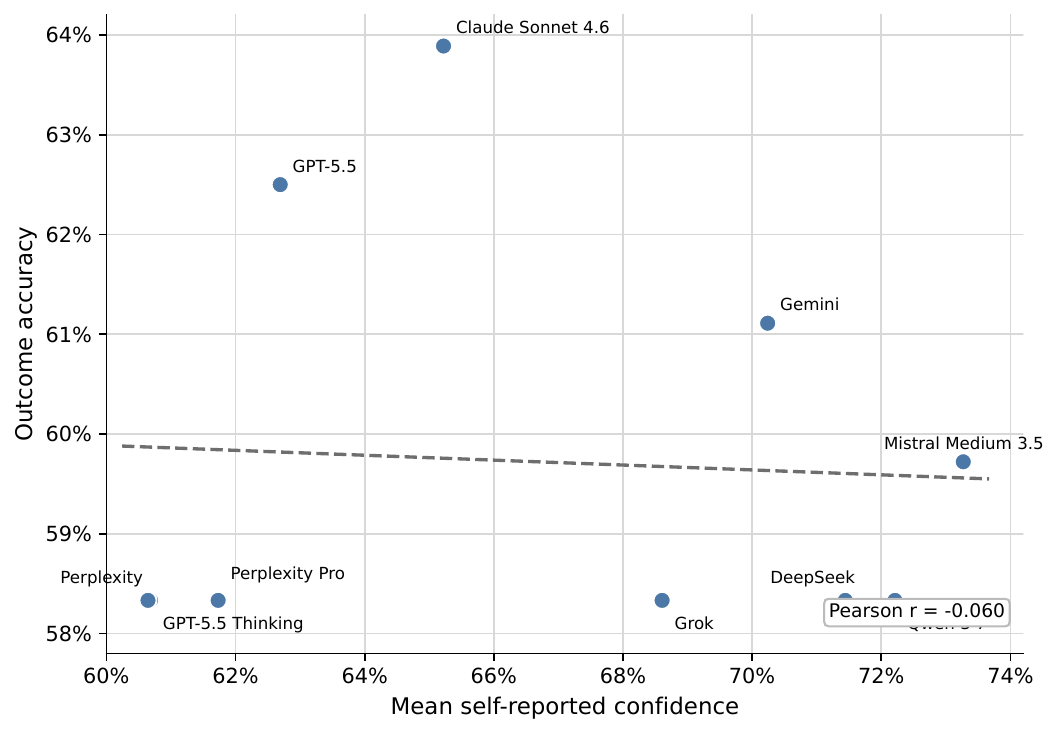}
\caption{Mean self-reported confidence versus group-stage outcome accuracy. The cross-model Pearson correlation is $r=-0.060$.}
\label{fig:confidence}
\end{figure}

Mean confidence ranged from 60.64\% to 73.27\%. Across models, confidence had almost no relationship with either total points ($r=-0.067$) or group-stage outcome accuracy ($r=-0.060$). As shown in Figure~\ref{fig:confidence}, the highest-confidence model, Mistral Medium 3.5, was not the most accurate, while the overall winner reported one of the lowest average confidence values.

These values should not be interpreted as a complete calibration analysis. Each model supplied one confidence value for its selected scoreline rather than a probability distribution over possible outcomes. The reported figures therefore describe expressed certainty, not calibrated predictive probabilities.

\section{Discussion}
\subsection{What did the benchmark measure?}
AI World Cup combines three related tasks. Models had to predict individual group-stage matches, build a consistent knockout bracket, and follow a demanding structured-output format. These tasks cannot be separated completely. A model may lose points because its bracket is internally inconsistent even when some of its football judgements are reasonable, while a correct champion prediction can offset fairly ordinary group-stage performance.

The knockout path created most of the separation between models. That is useful when the goal is to judge an end-to-end tournament forecast, but it also limits what can be inferred from the headline ranking. Claude performed best on match outcomes, Qwen and Perplexity Pro produced the most exact scores, and GPT-5.5 Thinking achieved the highest path-weighted total. The benchmark therefore has several leaders, depending on which forecasting skill matters most.

\subsection{Reasoning mode and within-provider comparison}
The two OpenAI-labelled submissions offer one useful, although very limited, within-provider comparison. GPT-5.5 Thinking scored 27 more total points than GPT-5.5 despite lower group-stage outcome accuracy (58.33\% versus 62.50\%) and a lower combined group-phase score (502 versus 525). The difference came from the tournament path: 242 versus 192 knockout points and a correct champion. The result is compatible with the idea that a longer reasoning process may help with bracket consistency, but one pair of submissions cannot establish that explanation. Repeated runs under controlled settings would be needed before drawing a causal conclusion.

The two Perplexity-labelled submissions illustrate a different pattern. Perplexity Pro tied for the highest group-stage points and exact-score accuracy, yet received no knockout points and finished below the non-Pro submission. Product tier alone therefore did not predict full-tournament performance in this sample.

\subsection{Sensitivity to the scoring weights}
The published points system intentionally rewards multiple levels of correctness, but its scale is not stage-normalised. A model can collect overlapping match-level bonuses, group-standing awards, and knockout progression points over teams and rounds. The realised benchmark contains two comparatively low-variance group components and one high-variance knockout component. Consequently, the knockout path can dominate rank differences even though group-stage and standing points together form most of the absolute score.

A useful robustness analysis would recompute the leaderboard under several reasonable alternatives: equal weighting of the group and knockout phases, normalisation by the maximum available score in each component, removal of the champion bonus, and scoring round qualification independently of exact bracket pairings. If the same models remain near the top, confidence in the ranking would increase. These alternatives should be fixed before the next tournament rather than chosen after the results are known.

\subsection{The value of a fixed pre-tournament forecast}
Other 2026 benchmarks refreshed forecasts before individual matches and could incorporate new information as the tournament progressed \citep{wang2026worldcuparena,ding2026wcagents,schroeder2026soccerarena}. AI World Cup deliberately did not. Its single pre-tournament submission tests whether a model can maintain a coherent long-range forecast from one fixed information state. The disadvantage is that the benchmark cannot tell whether later errors arose from weak reasoning or simply from information that became outdated after the prediction was made.

\section{Comparison with Contemporary Benchmarks}
Table~\ref{tab:relatedcomparison} summarises the complementary designs. AI World Cup is the only one among these studies centred on one complete pre-tournament bracket from a broad collection of consumer-accessible assistants. The other benchmarks are stronger on probabilistic evaluation, repeated temporal snapshots, market baselines, or detailed event prediction.

\begin{table}[H]
\centering
\caption{Comparison with contemporaneous LLM football forecasting benchmarks.}
\label{tab:relatedcomparison}
\resizebox{\textwidth}{!}{%
\begin{tabular}{lllll}
\toprule
Benchmark & Forecast timing & Systems & Primary outputs & Distinctive feature \\
\midrule
AI World Cup & Once, pre-tournament & 10 assistants & Scores, outcomes, full bracket, confidence & Accessible manual full-bracket comparison \\
WorldCupArena & Before each match & 13 systems & Scores, events, players, statistics & Fine-grained tasks and human/market baselines \\
WC2026-Agents & Before each match & 4 agents + market & 1X2 probabilities, bets, reflections & Search--act--reflect agents and economic baseline \\
LLM-SoccerArena & Multiple horizons & 7 LLMs & Match probabilities and tournament questions & Factorial prompt/tool/horizon design \\
\bottomrule
\end{tabular}}
\end{table}

Taken together, these benchmarks are better viewed as complementary than as competitors. A full-tournament forecast tests long-range consistency, per-match probabilities support calibration analysis, detailed event forecasts test evidence synthesis, and market comparisons provide a strong practical baseline.

\section{Threats to Validity and Limitations}
\paragraph{Small and non-random participant set.} The benchmark contains ten submitted configurations, not a statistically representative sample of LLMs. Provider counts are unbalanced, and model labels may refer to changing consumer products.

\paragraph{Single run per model.} Stochastic assistants can produce different brackets across sessions. Without repeated independent samples, observed differences combine model capability with sampling variation.

\paragraph{Information-access uncertainty.} The protocol recommends disabling web search and separating search-enabled assistants, but consumer interfaces can use hidden retrieval or tools. Access conditions are difficult to verify perfectly in a manual experiment.

\paragraph{Long-context and schema effects.} The full-tournament prompt is long, and models may differ in context handling or structured-output reliability. The result therefore measures the complete assistant workflow, not only latent football knowledge.

\paragraph{Composite-score subjectivity.} The point weights encode design choices. The final ranking was especially sensitive to knockout progression and the realised bracket.

\paragraph{Simplified standing reconstruction.} Actual group standings were reconstructed from points, goal difference, goals scored, and an alphabetical fallback. This is transparent and deterministic, but it may not reproduce every official tournament tiebreak procedure in rare tied cases. Future releases should encode the complete official hierarchy and publish tie-resolution tests.

\paragraph{Confidence is not a probability distribution.} Mean confidence cannot support full calibration analysis, and averages can hide match-level overconfidence or underconfidence.

\paragraph{No external baseline in the original leaderboard.} The benchmark did not include bookmaker probabilities, Elo-style statistical models, simple favourites, or human forecasters. It therefore compares submitted assistants with one another, not with the strongest available forecasting baseline.

\paragraph{Mutable publication surfaces.} The GitHub README, generated JSON, and deployed single-page website are separate publication surfaces and may be refreshed or cached at different times. The immutable commit and bundled result JSON are therefore the authoritative provenance for this paper. A tagged release and checksum manifest would further reduce ambiguity.

\paragraph{Limited public aggregation for bracket diagnostics.} The frozen leaderboard JSON preserves aggregate scores and champion predictions but not the full round-by-round bracket predictions in a compact analysis-ready table. This limits the depth of consensus visualisation that can be reproduced directly from the archived aggregate bundle and motivates releasing an explicit stage-advancement matrix in future versions.

\section{Recommendations for Using LLMs in Football and Tournament Prediction}
The findings from this study suggest several practical principles for researchers and practitioners who use LLMs to predict football matches or complete tournaments. These recommendations apply beyond the AI World Cup project and are intended to improve validity, transparency, and interpretability in future forecasting studies.

\subsection{Pre-register the forecasting protocol}
Before collecting predictions, define and publish the information snapshot, prompt, output schema, model-access conditions, scoring rules, and evaluation plan. Predictions should be timestamped and preserved in an immutable form so that they cannot be altered after outcomes become known. Versioned releases and checksums are useful for maintaining a clear audit trail.

\subsection{Use repeated independent forecasts}
A single response may be strongly affected by sampling variation, interface settings, or hidden system updates. Future studies should collect several independent forecasts from each model under the same conditions and report the mean, variability, and rank distribution for every metric. Repeated runs also make it possible to distinguish systematic model behaviour from one unusually successful bracket.

\subsection{Request probabilistic predictions}
For each match, models should provide a probability vector $(p_H,p_D,p_A)$ for home win, draw, and away win, with $p_H+p_D+p_A=1$. These forecasts can be evaluated using proper scoring rules such as the Brier score and logarithmic score \citep{brier1950verification,gneiting2007strictly}. If scorelines are required, models should provide either a probability distribution over plausible scores or the parameters of a goal model. Reliability diagrams may complement proper scores when enough predictions are available.

\subsection{Separate match-level and tournament-level performance}
Football prediction studies should report local and global performance separately. At minimum, results should distinguish:
\begin{enumerate}
    \item individual match-outcome prediction;
    \item exact-score or goal-distribution prediction;
    \item group-ranking and qualification prediction;
    \item knockout-stage progression; and
    \item final champion or complete-bracket prediction.
\end{enumerate}
A composite score can still be useful, but it should not replace the component metrics. When components have very different ranges or variances, they should be normalised before aggregation.

\subsection{Avoid overly brittle bracket scoring}
Exact brackets are path-dependent: one early error can make several later predictions impossible to score correctly. Tournament evaluations should therefore also award points for predicting whether a team reaches each round, regardless of the exact opponent path. Marginal probabilities of reaching the Round of 16, quarter-final, semi-final, final, and championship provide a more robust view of tournament expectations.

A released team-by-stage prediction matrix would also support consensus analysis and allow readers to compare how models differed in their expectations about each team's progression.

\subsection{Control and document information access}
Models with access to live web search, external tools, or autonomous agents should not be compared directly with models restricted to a fixed dataset unless the access conditions are clearly separated. Studies should record the prompt, model label, date, interface, enabled tools, cited sources, and tool traces where available. Distinct tracks for fixed-data, web-enabled, and agentic systems are preferable to a single mixed leaderboard.

\subsection{Compare against strong non-LLM baselines}
LLM forecasts should be evaluated alongside bookmaker-implied probabilities after margin removal, Elo-based systems, expected-goals models, ranking heuristics, host or favourite baselines, and human crowd forecasts. Contemporary work indicates that betting markets remain difficult for LLM-based systems to outperform consistently \citep{ding2026wcagents,wang2026worldcuparena}.

\subsection{Validate the evaluation pipeline}
Automated checks should verify fixture completeness, prediction counts, score decomposition, rank ordering, and reproducible regeneration of every table and figure. The pipeline should fail visibly when expected predictions or leaderboard entries are missing. These checks are especially important for long tournaments, where a small data or export error can affect many reported results.

\section{Conclusion}
AI World Cup shows that the informal predictions produced by consumer LLMs can be turned into a structured and auditable forecasting experiment. Under its completed three-component scoring system, GPT-5.5 Thinking ranked first with 744 points and was the only submission to predict champion Spain. The final ranking was strongly shaped by knockout progression: group-stage match and standing scores were comparatively clustered, while knockout points varied from zero to 242. Match-level metrics produced a different leader, with Claude Sonnet 4.6 recording the highest outcome accuracy and Qwen 3.7 and Perplexity Pro recording the most exact scores.

The final table should not be read as proof that one model simply ``understands football'' better than all others. A narrower conclusion is justified: under this particular scoring rule, one submitted configuration produced the best realised tournament path, while other models were better at predicting individual outcomes or exact scores. The lack of any clear relationship between confidence and accuracy also shows why self-reported certainty should not be treated as calibrated probability.

The most valuable outcome of the project is the transparent protocol and the preserved raw predictions. A future edition with immutable releases, repeated runs, probabilistic forecasts, proper scoring rules, stage-normalised reporting, external baselines, and stronger integrity checks could provide a more rigorous test of long-horizon LLM forecasting, both in sport and in other domains.

\section*{Data and Code Availability}
The benchmark code, protocol, prompt generation workflow, raw model responses, fixtures, and website are available through the project repository identified in the footnote in Section~1. The numerical results in this manuscript are frozen to commit \texttt{f83ea90}, which incorporates final group-standing scoring and generated website data for ten models, 1,040 predictions, and 104 fixtures. The source package accompanying this paper includes the exact leaderboard JSON used for all reported calculations.

\section*{Ethical and Responsible Use Statement}
This project evaluates forecasting systems for research and public benchmarking. It is independent of FIFA and is not betting or financial advice. Forecasts should not be used as guarantees of sporting outcomes.

\section*{Acknowledgements}
The authors thank the developers and providers of the evaluated assistants, together with the open-source communities whose tools and data supported the project. ChatGPT was used during manuscript editing and the preparation of project documentation; the authors reviewed and verified the final text and all reported results.

\bibliographystyle{plainnat}
\bibliography{references}

@article{wang2026worldcuparena,
  title        = {{WorldCupArena}: Fine-Grained Evaluation of Language Models and Deep-Research Agents on Football Forecasting},
  author       = {Wang, Zhaokai and Gui, Tianlin and Rao, Jiayuan and Di, Shangzhe and Tang, Yihong and Liang, Dingli},
  journal      = {arXiv preprint arXiv:2607.18084},
  year         = {2026},
  eprint       = {2607.18084},
  archivePrefix= {arXiv},
  primaryClass = {cs.CL}
}

@article{ding2026wcagents,
  title        = {FIFA World Cup 2026 as a Contamination-Free Benchmark for LLM Forecasting Agents: Four Models, a Bookmaker, and 104 Matches},
  author       = {Ding, Jiacheng and Guo, Cong and Xu, Jason},
  journal      = {arXiv preprint arXiv:2607.17765},
  year         = {2026},
  eprint       = {2607.17765},
  archivePrefix= {arXiv},
  primaryClass = {cs.CL}
}

@article{schroeder2026soccerarena,
  title        = {{LLM-SoccerArena}: Benchmarking LLMs on Real-World Predictions in Sports},
  author       = {Schröder, Jonas and Schweisthal, Jonas and Müller, Oliver and Weinmann, Markus and Feuerriegel, Stefan},
  journal      = {arXiv preprint arXiv:2607.24573},
  year         = {2026},
  eprint       = {2607.24573},
  archivePrefix= {arXiv},
  primaryClass = {cs.CL}
}

@article{brier1950verification,
  title   = {Verification of Forecasts Expressed in Terms of Probability},
  author  = {Brier, Glenn W.},
  journal = {Monthly Weather Review},
  volume  = {78},
  number  = {1},
  pages   = {1--3},
  year    = {1950}
}

@article{gneiting2007strictly,
  title   = {Strictly Proper Scoring Rules, Prediction, and Estimation},
  author  = {Gneiting, Tilmann and Raftery, Adrian E.},
  journal = {Journal of the American Statistical Association},
  volume  = {102},
  number  = {477},
  pages   = {359--378},
  year    = {2007}
}

@article{tetlock2014superforecasting,
  title   = {Forecasting Tournaments: Tools for Increasing Transparency and Improving the Quality of Debate},
  author  = {Tetlock, Philip E. and Mellers, Barbara A. and Rohrbaugh, Nick and Chen, Eva},
  journal = {Current Directions in Psychological Science},
  volume  = {23},
  number  = {4},
  pages   = {290--295},
  year    = {2014}
}

\appendix
\setcounter{table}{0}
\renewcommand{\thetable}{A\arabic{table}}
\section{Detailed Scoring Tables}
The following tables provide the complete point allocations used in the three scoring components described in the scoring methodology.

\begin{table}[H]
\centering
\caption{Cumulative scoring for each group-stage match.}
\label{tab:matchscoring}
\begin{tabular}{lr}
\toprule
Criterion & Points \\
\midrule
Exact home and away score & 5 \\
Correct outcome (home win, draw, or away win) & 3 \\
Correct winning team for a non-draw & 2 \\
Correct goal difference & 1 \\
\bottomrule
\end{tabular}
\end{table}

\begin{table}[H]
\centering
\caption{Cumulative scoring for each predicted group standing.}
\label{tab:standingscoring}
\begin{tabular}{lr}
\toprule
Criterion & Points \\
\midrule
Correct group winner & 5 \\
Correct set of the top two teams & 5 \\
Each correctly predicted qualifier from the group & 3 per team \\
Each team placed at its exact final rank & 2 per team \\
\bottomrule
\end{tabular}
\end{table}

\begin{table}[H]
\centering
\caption{Knockout progression and final-placement scoring.}
\label{tab:knockoutscoring}
\begin{tabular}{lr@{\hspace{2.5em}}lr}
\toprule
Progression criterion & Points & Final placement & Points \\
\midrule
Each correct Round-of-32 team & 2 & Correct champion & 20 \\
Each correct Round-of-16 team & 4 & Correct runner-up & 10 \\
Each correct quarter-finalist & 6 & Correct third place & 8 \\
Each correct semi-finalist & 8 & Correct fourth place & 5 \\
Each correct finalist & 12 & & \\
\bottomrule
\end{tabular}
\end{table}

\section{Reproducibility Checklist}
\begin{table}[H]
\centering
\caption{Minimum artefacts required to reproduce the reported evaluation.}
\begin{tabularx}{\textwidth}{lX}
\toprule
Artefact & Requirement \\
\midrule
Data freeze & Exact tournament snapshot and fixture-result file \\
Prompt & Full generated prompt and prompt-version identifier \\
Submissions & Raw response for every model, preserved without editing \\
Metadata & Model label, provider, access mode, timestamp, enabled tools \\
Parser & Versioned schema, validation rules, and import logs \\
Scorer & Versioned scoring implementation and unit tests \\
Release & Immutable commit/tag and checksums for all result artefacts \\
Environment & Python and dependency versions plus clean-build instructions \\
\bottomrule
\end{tabularx}
\end{table}

\section{Descriptive Statistics}
Across the ten submissions, mean total score was 618.3 (sample standard deviation 86.65), mean group-stage match score was 258.9 (8.65), mean group-standing score was 248.6 (12.20), and mean knockout score was 110.8 (88.65). The combined pre-knockout score averaged 507.5 points (14.62) and ranged from 491 to 530. Mean outcome accuracy was 59.72\%, mean exact-score accuracy was 10.14\%, and mean reported confidence was 66.67\%. These summary values are descriptive and should not be interpreted as estimates over a broader model population.

\begin{figure}[t]
\centering
\includegraphics[width=0.78\textwidth]{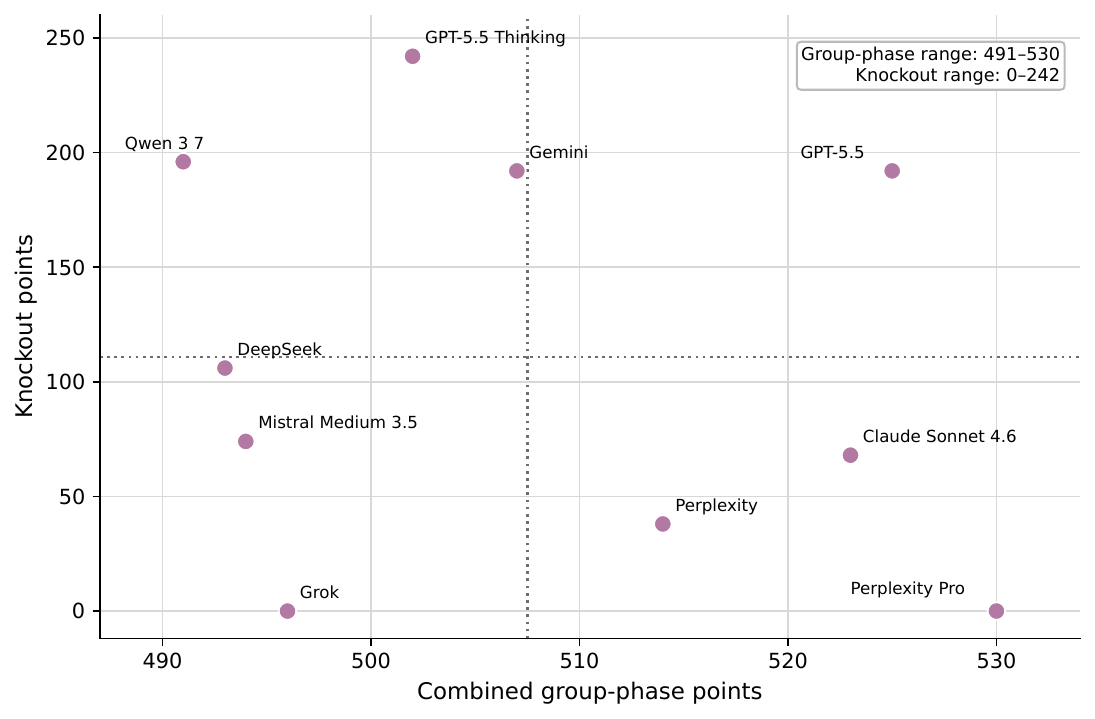}
\caption{Combined group-phase points (group-stage matches plus group standings) versus knockout points. The pre-knockout spread is small compared with the knockout spread.}
\end{figure}

\end{document}